\pdfoutput=1

\documentclass[11pt]{article}

\usepackage[final]{acl}

\usepackage{times}
\usepackage{latexsym}
\usepackage{amsmath}
\usepackage[T1]{fontenc}
\usepackage{graphicx}
\usepackage{CJKutf8}
\usepackage{amsthm,amsmath,amssymb}
\usepackage{subfigure}
\usepackage{mathrsfs}
\usepackage{booktabs}
\usepackage{enumerate}
\usepackage{bbding}
\usepackage{multirow} 
\usepackage{cite}
\usepackage{utfsym}
\usepackage{times}
\usepackage{latexsym}
\usepackage[T1]{fontenc}
\usepackage[utf8]{inputenc}
\usepackage{microtype}
\usepackage{inconsolata}
\usepackage{mathrsfs}

\usepackage[utf8]{inputenc}

\usepackage{microtype}

\usepackage{inconsolata}

\usepackage{graphicx}

\newcommand{\ie}{i.\,e.\@}
\title{See Detail Say Clear: Towards Brain CT Report Generation via Pathological Clue-driven Representation Learning}

\author{
 \textbf{Chengxin Zheng\textsuperscript{1}},
 \textbf{Junzhong Ji\textsuperscript{1}},
 \textbf{Yanzhao Shi\textsuperscript{1}},
 \textbf{Xiaodan Zhang\textsuperscript{1,*}},
 \textbf{Liangqiong Qu\textsuperscript{2}}
\\
\\
 \textsuperscript{1}College of Computer Science, Beijing University of Technology, Beijing, China \\
 \textsuperscript{2}Department of Statistics and Actuarial Science, School of Computing and Data Science,\\University of Hong Kong, Hong Kong, China
\\
 \small{
   \textbf{Correspondence:} \href{mailto:zhangxiaodan@bjut.edu.cn}{zhangxiaodan@bjut.edu.cn}
 }
}

\begin{document}
\maketitle
\begin{abstract}

Brain CT report generation is significant to aid physicians in diagnosing cranial diseases.
Recent studies concentrate on handling the consistency between visual and textual pathological features to improve the coherence of report.
However, there exist some challenges: 
1) \textbf{Redundant visual representing}: 
Massive irrelevant areas in 3D scans distract models from representing salient visual contexts.
2) \textbf{Shifted semantic representing}: Limited medical corpus causes difficulties for models to transfer the learned textual representations to generative layers. 
This study introduces a Pathological Clue-driven Representation Learning (PCRL) model to build cross-modal representations based on pathological clues and naturally adapt them for accurate report generation.
Specifically, we construct pathological clues from perspectives of segmented regions, pathological entities, and report themes, to fully grasp visual pathological patterns and learn cross-modal feature representations. To adapt the representations for the text generation task, we bridge the gap between representation learning and report generation by using a unified large language model (LLM) with task-tailored instructions. These crafted instructions enable the LLM to be flexibly fine-tuned across tasks and smoothly transfer the semantic representation for report generation.
Experiments demonstrate that our method outperforms previous methods and achieves SoTA performance.
Our code is available at 
\href{https://github.com/Chauncey-Jheng/PCRL-MRG}{https://github.com/Chauncey-Jheng/PCRL-MRG}.

\end{abstract}

\section{Introduction}

\let\thefootnote\relax\footnotetext{*Corresponding Author}
Brain computed tomography (CT) imaging is essential for diagnosing various cranial diseases, including cerebral infarction and hemorrhage.
However, it is time-consuming and error-prone for radiologists to manually interpret medical findings from these scans and write reports.
Automated report generation systems are designed to boost efficiency, reduce the workload for radiologists, and optimize resources in busy clinical scenarios.

\begin{figure}
\centering  
\includegraphics[width=0.49\textwidth]{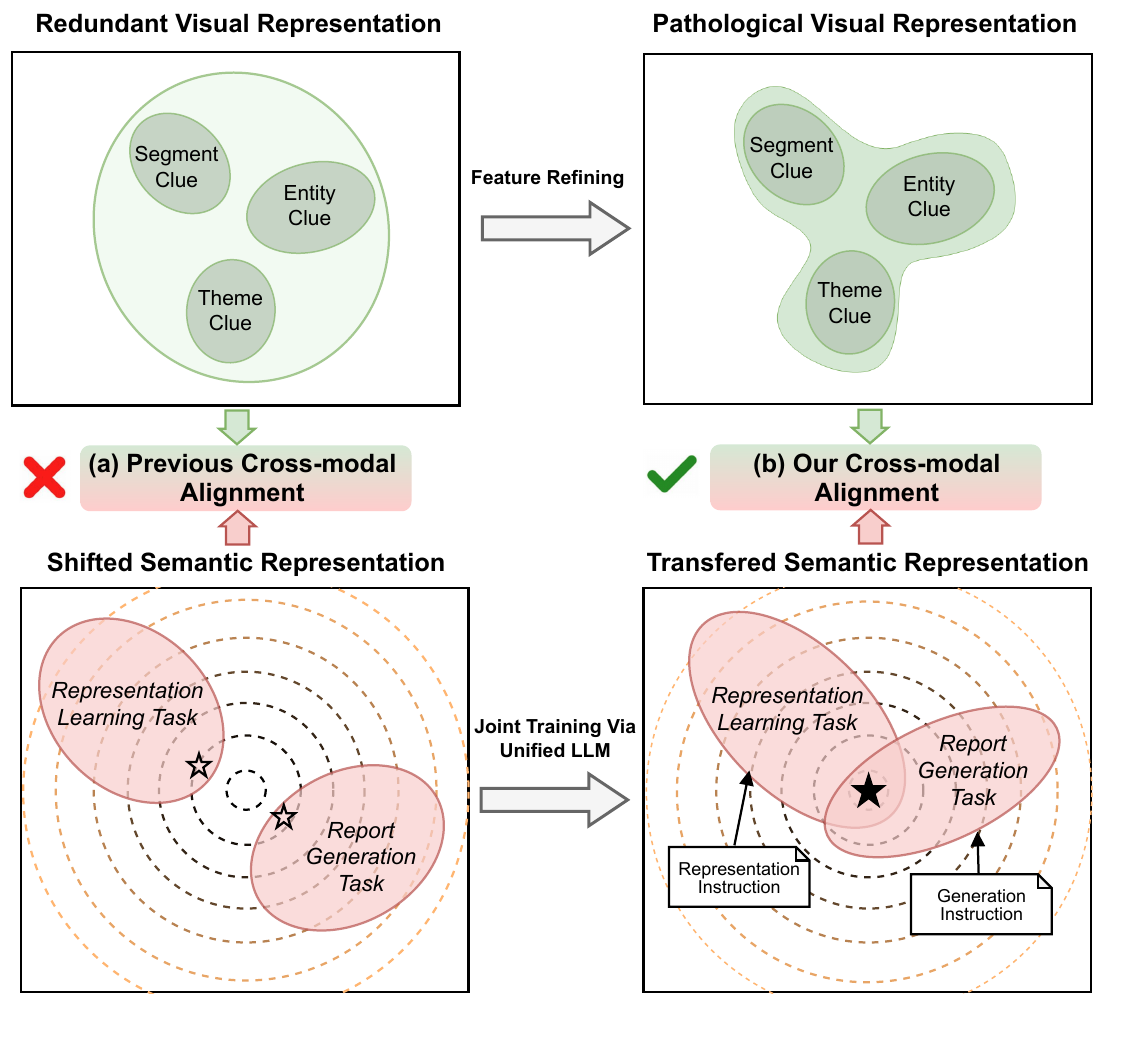}
\caption{
Comparison of different cross-modal alignment paradigms.
\textbf{(a) Previous way}: The origin CT images contain extraneous information unrelated to diagnosis, and separate training for two semantic tasks makes it challenging to find a shared optimal solution, resulting in inadequate cross-modal alignment. 
\textbf{(b) Our way}: The visual representation is refined to concentrate on pathological clues, and employ joint training with task-tailored instructions via unified LLM to find a transferable representation, leading to better adaption for report generation.
} \label{fig1}
\end{figure}

With the advancement of deep neural networks and their successful application in image captioning tasks~\citep{vinyals2015show,xu2015show}, medical report generation (MRG) has gained more attention.
Unlike the short sentences of traditional image captioning, MRG aims to generate lengthy and precise reports. 
To achieve this, various cross-modal alignment methods are required to ensure the consistency between visual and textual information~\citep{Shi2023Granularity}, including attention mechanisms~\citep{jing2018automatic, wang2018tienet}, memory mechanisms~\citep{chen2020generating,chen2022cross}, and knowledge graphs~\citep{li2019knowledge,li2023dynamic}.

Recently, learning representations via visual-textual contrastive learning~\citep{Shi2023Granularity, Shi2024Prior} or using pre-trained large language models (LLMs)~\citep{Thawakar2023XrayGPT} to strength representations are also proven to be effective.

However, as shown in Figure~\ref{fig1}(a), learning cross-modal correspondences is still challenging in current methods due to the following concerns:
1)~\textbf{Redundant visual representing.}
Different from chest X-ray data, 3D brain CT scans contain extensive redundant information, e.g. background and insignificant areas.
With the lack of human-crafted boxes to locate pathology regions, models struggle to capture and interpret the visual pathology patterns for generating reports.
Although current advanced models use semantic prior knowledge or medical prompts~\citep{Jin2024PromptMRG, Bu2024Dynamic} to automatically learn the salient visual areas, this may introduce noise and unstable representation and cause severe hallucinations in generated texts.
2) \textbf{Shifted semantic representing.}
Compared to natural corpus, limited brain CT report corpus is insufficient to transfer the pathological semantic representations learned by represent learning layers (e.g., contrastive learning layer) to the language model~\citep{huh2024platonic}, since the direct weight-sharing~\citep{Shi2023Granularity, Shi2024Prior} is prone to degrade the coherence of generated diagnostic sentences.
Thus, how to uniformly represent cross-modal pathological features and adapt them to report generation still remains an open question.

In this paper, we propose a Pathological Clue-driven Representation Learning (PCRL) model to seamlessly build cross-modal representations based on diverse pathological clues and transfer them for generating accurate brain CT reports. 
Specifically, we extract pathological clues from perspectives of segmented regions, pathological entities, and report themes to depict clinical scenarios.
Segmented region clues are automatically generated and filtered by given pathology prompts, enabling the visual encoder to grasp visual pathological patterns.
Meanwhile, entity and theme clues are respectively extracted by detailed findings and full-text reports, to handle the enriched visual-textual alignment and build cross-modal pathology representations.

Besides, to adapt the learned representations for the text generation task, we bridge the gap between representation learning and report generation by employing a unified large language model (LLM) with task-tailored instructions, which has proven to be more effective than conventional decoders by using appropriate tokens to seamlessly connect different tasks~\citep{yu2023language}.
As shown in Figure \ref{fig1}(b), We craft a representation instruction to prompt the LLM to produce high-level pathological semantic features for cross-modal alignment, 
and a generation instruction to prompt LLM to generate accurate brain CT reports based on the learned representations.

Our main contributions can be summarized as:
\begin{enumerate}

\item We propose a novel framework to seamlessly learn visual-textual representations from perspectives of diverse pathological clues and leverage them for enhancing the quality of generated brain CT reports.

\item We, for the first time, design a new paradigm to effectively transfer the learned pathological representations for report generation using a unified LLM prompted by task-tailored instructions.

\item We validate the model capabilities on the open-source CTRG-Brain dataset. Experimental results show that our model 
achieves remarkable performance in generating brain CT reports.

\end{enumerate}

\begin{figure*}
    \centering
    \includegraphics[width=1.00\linewidth]{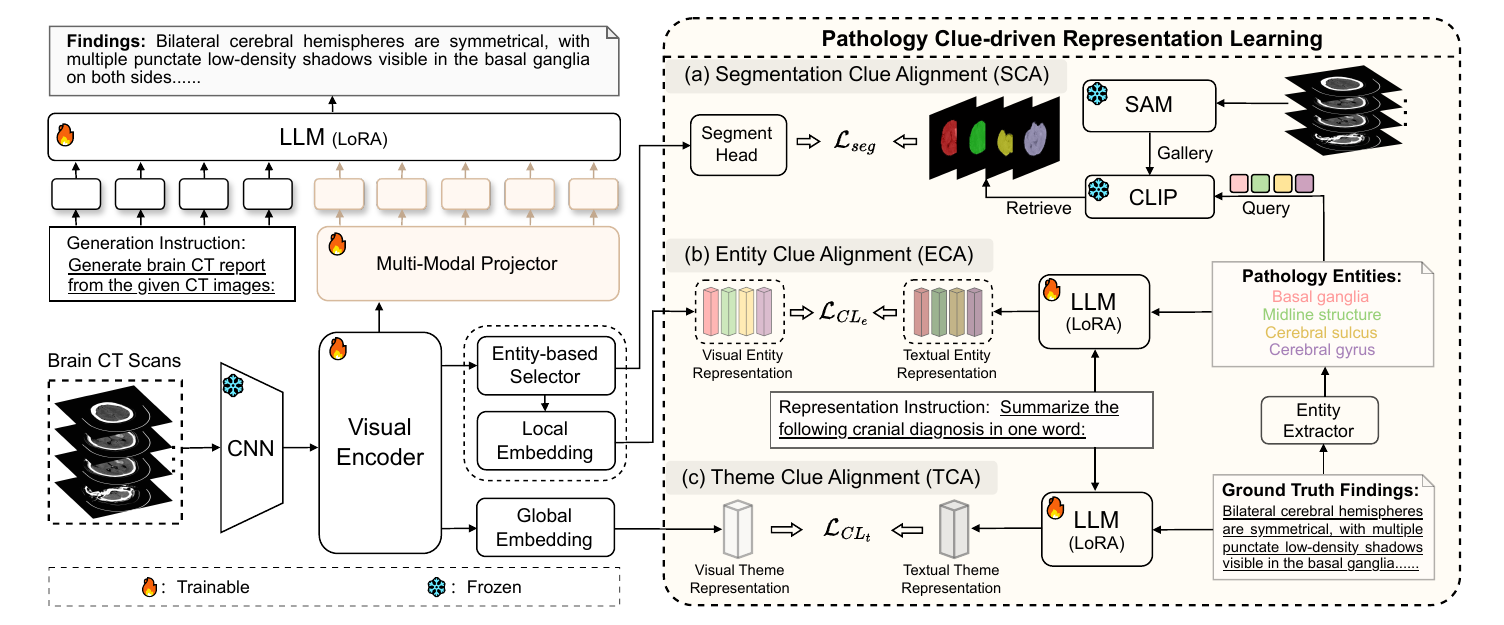}
    \caption{The overall framework of our method, which mainly consists of an image encoder and a text decoder for brain CT report generation (left). The pathological clue-driven representation learning (right) is proposed to guide the encoder and decoder for more fine-grained representation by three alignment modules: (a) segmentation clue alignment (SCA), (b) entity clue alignment (ECA), and (c) theme clue alignment (TCA).}
    \label{fig:SDSC_main}
\end{figure*}

\section{Related Work}

\subsection{Medical Report Generation}
The advancements in image captioning techniques have spurred the development of a range of radiology report generation methods~\citep{jing2018automatic, chen2020generating, chen2022cross, li2019knowledge, Shi2023Granularity,zhang2023weakly,li2023dynamic,Shi2024Prior,shen2024ghcl}.
To exploit the more effective parts of medical images,
\citet{jing2018automatic} first propose a co-attention mechanism to associate images with disease tags and improve the accuracy of generated reports;
\citet{chen2022cross} design a cross-modal memory matrix to learn high-level vision-language correspondence for enhancing report generation;
\citet{liu2021contrastive} used contrastive attention to captured the difference of abnormal and normal samples.
To address the semantic bias of limited medical report corpus,
\citet{jing2020show} exploited the textual structure information of reports;
\citet{liu2021competence} used curriculum learning to alleviate the data bias of limited medical data corpus.
The retrieval-based \citet{li2018hybrid,liu2021exploring} and prior expert knowledge based methods \citep{zhang2020radiology, yang2022knowledge, you2021aligntransformer} also shows 
effectiveness to alleviate the inaccuracy of generated report caused by limited training data.
To build more stable relations between multi-modal features, contrastive learning~\citep{radford2021learning} acts as an unsupervised method, and is proven to be effective for learning cross-modal representations for MRG tasks~\citep{li2023dynamic, Shi2024Prior}.

Recently, LLM-based MRG methods~\citep{Jin2024PromptMRG, Chen2024Dia} harness the semantic processing capabilities of LLMs to enhance the coherence of generated medical reports.
However, the current bottleneck of LLM is to interpret sparse visual patterns of pathologies, which causes severe hallucinations and degrades the trustworthiness of clinical diagnosis.
To solve this issue, we propose to augment LLM with representation learning guided by pathology clues. By aligning cross-modal pathological features and transferring the learned representations to the LLM-based report decoder, our model can generate accurate and coherent brain CT reports.

\subsection{Pre-trained Large Model}
In recent years, pre-trained large models have achieved significant breakthroughs in both natural language processing (NLP) and computer vision (CV). These models leverage extensive datasets for pre-training, allowing them to perform exceptionally well in downstream tasks. 
The Segment Anything Model (SAM) \citet{kirillov2023segment} is an innovative large vision model in the field of image segmentation. 
SAM’s strengths lie in its extensive training on a large-scale dataset, enhancing its versatility and enabling real-time generation of high-quality segmentation masks across diverse tasks and image distributions.
MedCLIP \citet{wang2022medclip} is a notable visual-language model in the medical domain. 
Pre-trained on decoupling medical images and texts, MedCLIP has shown SoTA performance on zero-shot image-text retrieval.
LLaMA3 is the next generation of state-of-the-art open-source LLM from \citet{meta2024introducing}. 
Trained on over 15 trillion tokens, seven times the data of its predecessor, LLaMA3 incorporates extensive multilingual and high-quality datasets. 
To improve the use of large models for assisting in medical report generation, we adopt SAM for region segmentation, MedCLIP for feature representation of pathological entities, and LLaMA for joint training to achieve fine-grained cross-modal alignment and improve medical report generation.

\section{Methodology}

As illustrated in Figure~\ref{fig:SDSC_main}, our framework consists of two branches: the brain CT Report Generation (RG) on the left and the Pathological Clue-driven Representation Learning (PCRL) on the right. The interaction between these branches is facilitated through the shared use of a visual encoder and a language model.

\subsection{Brain CT Report Generation}
In this branch, the input comprises a set of CT scan images \( I = \{i_1, \ldots, i_N\} \), where \( N \) represents the number of CT images in each sample. The output is the corresponding brain CT findings report \( Y = \{y_1, \ldots, y_M\} \), with \( M \) representing the number of tokens.
We adopt the encoder-decoder architecture for report generation. 
First, we use ResNet101~\citep{he2016deep} to extract grid features \( G = \{g_1, \ldots, g_N\} \in \mathbf{R}^{N\times H \times d}(N = 24,H = 196, d = 2048)\) from \( I \). Then, we embed these features using a visual encoder to obtain visual features \( V \). 
These visual features \( V \) are passed through a multi-modal projector to obtain the visual embedding tokens \( H_v \in \mathbf{R}^{N \times d_w }(d_w = 4096)\), which are aligned with the word embedding space of a large language model. 
Finally, these tokens are concatenated with the instruction token \( H_q \) and fed into the large language model. 
We train the parameters \( \theta \) by minimizing the cross-entropy loss, which can be represented by the following formula:
\begin{equation}
\mathcal{L}_g = - \sum_{t=1}^M \log P(y_t \mid y_{1:t-1}, H_v, H_q; \theta)
\end{equation}
where, $P(y_t|*)$ denotes the probability conditioned on \( H_v, H_q \), and the embeddings of previous words \( y_1, y_2, \ldots, y_{t-1} \).

\subsection{Pathological Clue-driven Representation Learning}
Learning fine-grained visual-text representations is critical for generating accurate reports~\citep{Shi2023Granularity}.
This branch mainly builds enriched representations by conducting multi-modal feature alignment based on pathological clues, including segmentation clue alignment (SCA), entity clues alignment (ECA), and theme clues alignment (TCA).




\subsubsection{Preparation of Pathological Clues
}
Pathological clues are collected in three perspectives for feature alignment.

\textbf{Pathology Theme Clues}:
Learning the structure of professional brain CT reports is essential for MRG models to satisfy human standards and produce reliable reports.
To this end, we propose to build theme clues by simply using the full-text report and whole images as global signals, which can be useful to enhance the overall quality of medical reports by TCA (illustrated in \ref{TCA}). 

\textbf{Pathology Entity Clues}:
To learn detailed information about pathology entities, we regard each single finding sentence in the report as an entity clue and extract the related multi-modal features in ECA~\ref{ECA}.
The construction of entities \( E = \{e_1,...,e_{N_e}\} \) (with \( N_e = 24 \)) is based on expert knowledge and the frequency of words in the training corpus. 

\textbf{Segment Clues}:
To enhance visual representations via detailed contour of pathologies, we propose to utilize the pre-trained segmentation model SAM~\citep{kirillov2023segment} to generate mask candidates and filter useful masks related to pathology entities.
First, we prompt SAM by covering each brain CT image with a grid of points and integrating it into image embeddings through average sampling.
In this way, the mask decoder in SAM is prompted to generate a gallery of candidate masks \(M_I = \{ m_1,...,m_{N_m}\}\).
To ensure the quality of masks, we also apply rule-based methods to filter out low-quality and duplicate masks with area size, stability scores, and IoU scores.
For each sample consisting of 24 images, we generate a corresponding series of masks and combine them to form the candidate \(Gallery = \{ M_{I_1},...,M_{I_N}\}\).

Then, to retrieve valuable masks related to the sample's pathological entities, we propose to use the MedCLIP~\citep{wang2022medclip} for text-prompted retrieval.
Based on expert knowledge, we divide the 24 images into eight-layer categories, each mapping a specific set of entities~\citep{Shi2024Prior}. 
Conversely, each entity also has its corresponding layers.
We extract the existing entities from the sample's report and their corresponding pathological descriptions \( D = \{d_1, ..., d_{N_d}\} \) (\( N_d \leq N_e \)), which serve as the \(Query\).
Next, we use the $d$-th description $Query_d$ to search for the most similar matching mask in $Gallery_d$, which is a subset of $Gallery$, consisting of all visual masks of the layers related to the existing entities.
The procedure can be represented as:
\begin{equation}
Retrieval_d = CLIP(Gallery_d, Query_d)
\end{equation}
where $Retrieval_d$ denotes the retrieved masks for SCA~\ref{SCA}.

\subsubsection{Segmentation Clue Alignment}
\label{SCA}
This module aims to learn the fine-grained visual representation based on the extracted segment masks.
First, we obtain the visual features $V_{D} = \{v_1,...v_{N_d}\} $ corresponding to the images containing the entities using a selector. 
Then, we input these features into a lightweight segmentation head to generate the corresponding foreground entity masks $S_{D} = \{s_1,...,s_{N_d}\} $, which serves as discriminated pathological information. 
We then align these generated masks with the retrieved segment masks $M_{D} = \{Retrieval_1,...,Retrival_{N_d}\} $ to learn detailed patterns. 
It is important to note that the size of the generated masks $S_{D}$ differs from the size of the retrieved SAM masks $M_{D}$. 
To address this, we first resize the SAM masks to match the size of the masks generated by our segmentation head. 
We finally calculate the following loss for aligning the two types of masks:
\begin{equation}
\mathcal{L}_{seg} = 1 - \frac{2\sum^{n}_{i=1}p_i y_i}{\sum^{n}_{i=1}p^2_i + \sum^{n}_{i=1}y^2_i}
\end{equation}
where $y_i$ is the pixel value of retrieved SAM masks and $p_i$ is the predicted pixel value.

In this way, the visual encoder can be effectively learned to focus on the areas of pathologies while reducing the influence of irrelevant visual features.


\subsubsection{Entity Clue Alignment}
\label{ECA}
To learn the cross-modal patterns of pathology entities, we extract visual and textual pathological features based on the entity clues.
First, we build a selector to obtain the visual features $V_{D} = \{v_1,...v_{N_d}\} (N_d < N)$, which corresponds to significant CT images that exist pathology entities.
We then apply global average pooling (GAP) to generalize $V_{D}$ and obtain representations of each significant CT image, denoted as $R_{v}^e = \{r_1,...r_{N_d}\}$.

Different from previous work~\citep{li2023dynamic} use an external language model (e.g. SciBert~\citep{Beltagy2019SciBERT}) to build textual representation for feature alignment, we propose to leverage a unified LLM to generate textual representation via tailored prompts. 
Inspired by~\citet{Wang2023ImprovingTE}, we crafted a representation instruction ``\textit{Summarize the following cranial diagnosis in one word:}'', which is denoted as $H_{q_r}$.
This prompt can effectively activate the summarization ability of LLM to generate corresponding text representations for pathological entity description in $D$.
We use the output generated by the final hidden layer of LLM as textual features for cross-modal alignment.

With the carefully extracted visual and textual representations of entities, we map them into the same dimension through an embedding layer.
The process can be represented by the following formula:
\begin{eqnarray}
    R_v^e &=& Linear_v^e(GAP(V_D)) \\
    R_w^e &=& Linear_t^e(LLM(H_{q_r}, D))
\end{eqnarray}
where $GAP$ denotes the global average pooling, $Linear_v$ represents the visual mapper, and $Linear_t$ represents the textual mapper.
We employ the symmetric InfoNCE \citep{oord2018representation} loss for visual-textual alignment:

\begin{align}
\mathcal{L}_{CL_e} = &- \frac{1}{2} \left(\alpha_v \sum_{i=1}^{N_d}\log \frac{\exp(\text{s}_{v}^{e}(i, i))}{\sum_{j=1, j\neq i}^{N_d} \exp(\text{s}_{v}^{e}(i, j))} \right. 
\notag \\
&+ \left. \alpha_w \sum_{i=1}^{N_d}\log \frac{\exp(\text{s}_{w}^{e}(i, i))}{\sum_{j=1, j\neq i}^{N_d} \exp(\text{s}_{w}^{e}(i, j))} \right)
\end{align}

where ${s}_{v}^{e}(i,j) = sim(R_{v_i}^e, R_{w_j}^e) / \tau$ and
${s}_{w}^{e}(i,j) = sim(R_{w_i}^e, R_{v_j}^e) / \tau$ denote the similarity between the visual representation \(R_v^e\) and the textual representation \(R_w^e\), \(\tau\) is a temperature parameter.
$\alpha_v$ and $\alpha_w$ are hyperparameters to balance the contrastive learning.
 
By aligning multi-modal entity clues, the model can grasp fine-grained visual-text representations to generate accurate diagnostic words.

\subsubsection{Theme Clue Alignment}
\label{TCA}
Theme clue alignment aims to equip the model with comprehensive skills to generate accurate style and structure of reports, thereby enhancing clinical reliability.
%
For global visual features, we use one-dimensional global max pooling (GMP) to represent the entire sample visually, denoted as $R^t_v$.
For the overall report of the sample, we generate textual representation $R^t_w$ by prompting LLM with the same representation instruction (see \ref{ECA}).
This process can be represented as:
\begin{eqnarray}
R_v^t &=& Linear_v^t(GMP(V)) \\
R_w^t &=& Linear_t^t(LLM(H_{q_s}, Y))
\end{eqnarray}
Similar to ECA, the loss of TCA can be formulated as:
\begin{align}
\mathcal{L}_{CL_t} = &- \frac{1}{2} \left(\alpha_v \sum_{i=1}^{N_b}\log \frac{\exp(\text{s}_{v}^{t}(i, i))}{\sum_{j=1, j\neq i}^{N_b} \exp(\text{s}_{v}^{t}(i, j))} \right. 
\notag \\
&+ \left. \alpha_w \sum_{i=1}^{N_b}\log \frac{\exp(\text{s}_{w}^{t}(i, i))}{\sum_{j=1, j\neq i}^{N_b} \exp(\text{s}_{w}^{t}(i, j))} \right)
\end{align}
where ${s}_{v}^{t}(i,j) = sim(R_{v_i}^t, R_{w_j}^t) / \tau$ and
${s}_{w}^{t}(i,j) = sim(R_{w_i}^t, R_{v_j}^t) / \tau$ denote the similarity between the visual representation \(R_v^e\) and the textual representation \(R_w^e\), \(\tau\) is a temperature parameter, $N_b$ is the batch size. Shared with ECA, $\alpha_v$ and $\alpha_w$ are set to balance the feature alignment.

\begin{table*}[t]
\centering
\begin{center}
\scalebox{0.9}{
\renewcommand\arraystretch{1}
\setlength{\tabcolsep}{6.8pt}{
\begin{tabular}{l|l|ccccccc|ccc}
\toprule
\multicolumn{1}{l|}{Methods}&
\multicolumn{1}{l|}{Decoder}&
\multicolumn{1}{c}{B1}&
\multicolumn{1}{c}{B2}&
\multicolumn{1}{c}{B3}&
\multicolumn{1}{c}{B4}&
\multicolumn{1}{c}{M}&
\multicolumn{1}{c}{RG}&
\multicolumn{1}{c|}{C}&
\multicolumn{1}{c}{F1}\\


\midrule

HRNN\citep{krause2017hierarchical}$^{\dagger}$&LSTM &42.3 & 28.3 & 21.2 & 17.1 & 27.2 & 39.3 & 20.9 
&  70.2    \\
Up-Down\citep{anderson2018bottom}$^{\dagger}$ &LSTM &45.8& 34.8& 28.5 & 24.4 & 31.6 & 42.5 & 27.3 
& 70.2    \\
WCL\citep{yan21weakly}$^{\dagger}$
&LSTM &49.5 & 36.5 & 29.4 & 25.1 & 31.3 & 42.8 & 33.3 
& 64.5  \\
R2Gen-CMN\citep{chen2022cross}$^{\dagger}$ &Transformer &49.1 & 40.0 & 34.4 & 30.1 & 29.9 & 48.6 & 84.2 
& 69.8 \\
XProNet\citep{wang2022cross}$^{\dagger}$
&Transformer &50.6 & 41.3 & 34.4 & 29.1 & 31.5 & 51.7 & 83.3 
& 70.1  \\
WGAM\citep{yang2021weakly}$^{\dagger}$
&LSTM & 49.4 & 36.7 & 29.6 & 25.4 & 32.0 & 42.4 & 31.9 
& 68.9  \\
PGCA\citep{Shi2023Granularity}$^{\dagger}$ &LSTM &50.2 & 37.8 & 30.7 & 26.5 &  32.5 & 43.0 & 34.0 
&  69.2\\
WGAM-HI\citep{zhang2023weakly}$^{\dagger}$ 
&LSTM  & 50.4 & 37.6 & 30.5 & 26.1 & 31.4 & 43.8 & 33.2 
& 67.4 \\
LLaVA-med\citep{li2023llava}$^{\dagger}$ 
&LLaMA3-8B &50.0 & 39.3 & 32.0 & 26.3 & 31.4 & 46.7 & 38.6
& 64.3 \\
HILT\citep{liu2024benchmarking}$^{\dagger}$ &LLaMA3-8B & 50.7 & 39.9 & 33.1 & 27.9 & 30.8 & 46.1 & 43.7
& 68.4 \\
PromptMRG\citep{Jin2024PromptMRG}$^{\dagger}$ 
&LLaMA3-8B &48.1 & 38.3 & 31.6 & 26.5 & 31.0 & 47.4 & 50.3
& 69.2 \\

\midrule
Ours (Jieba tokenizer)
&LLaMA3-8B
& 51.5 & 42.0 & 35.7 & 30.9 & 31.4 & 49.0 &  80.0 
& {} \\
Ours (LLaMA tokenizer)
&LLaMA3-8B
&\bfseries 62.0 & \bfseries 54.7 & \bfseries 49.4 & \bfseries 45.3 & \bfseries33.1 & \bfseries57.7 &  \bfseries 96.4 
& {\bfseries 70.6} \\
\bottomrule
\end{tabular}
}
}
\end{center}
\caption{The performance of our PCRL compared with previous state-of-the-art models on the Brain CT report generation dataset \textit{CTRG-Brain}. The best results are highlighted in bold.
${\dagger}$ denotes the re-implementation results.
}
\label{tb1:table1}
\end{table*} 

\subsection{Joint Training}
In the training stage, we jointly train the RG branch and the PCRL branch to maximize the utilization of multi-granularity visual-text representations.
Instead of using separate modules to learn representation and generate medical reports~\citep{li2023dynamic}, we use a unified LLM to bridge the gap between representation learning and report generation via two task-tailored instructions, i.e., representation instruction and generation instruction.
This can transfer the representations learned by the PCRL branch to optimize the RG branch effectively, thereby generating accurate reports.

Our final loss contains the above-mentioned losses, which can be formulated as:
\begin{equation}
    \mathcal{L} = \mathcal{L}_{g} + \mathcal{L}_{seg} + \mathcal{L}_{CL_e} + \mathcal{L}_{CL_t}
\end{equation}

\section{Experiments}

\subsection{Dataset}
We validate the performance of our model using the \textit{CTRG-Brain} \citep{tang2024work} dataset. This dataset comprises 6,000 samples, containing a total of 160,336 CT images and 6,000 Chinese medical reports. Following the mainstream division method~\citep{Shi2023Granularity,zhang2023weakly}, we split the dataset into a training set, a validation set, and a test set in a 7:1:2 ratio. 
For consistent processing, we divide the brain CT image samples into 8 layers based on expert knowledge, with each layer containing 3 continuous CT images, assigning each sample with 24 CT images.

\subsection{Evaluation Metrics}
We chose Natural Language Generation (NLG) metrics and Clinical Evaluation (CE) metrics to evaluate our model's performance. NLG metrics include BLEU \citep{papineni2002bleu}, METEOR \citep{banerjeemeteor}, ROUGE-L \citep{lin2004rouge}, and CIDEr \citep{vedantam2015cider}, denoted as B1, B2, B3, B4, M, RG, and C respectively. 
To measure the pathological accuracy, we use 24 keywords summarized by experienced radiologists to calculate the Clinical Evaluation (CE) metric\citep{Shi2023Granularity,zhang2023weakly}, \ie, F1 score, which is the harmonic mean of the precision and recall.

\subsection{Implementation Details}

We reshape the size of image to 512x512 pixels and used a ResNet101 to extract image features, which is pre-trained on the ImageNet dataset and fine-tuned on the CQ500 dataset~\citep{chilamkurthy2018development}. For our large language model, we utilize LLaMA3-8B \citep{meta2024introducing}, which is quantized to 4-bit, and use the LoRa \citep{hu2021lora} for parameter-efficient fine-tuning. 
The overall trainable parameter quantity of our model is 229.9M, with 3.4M parameters in LLM (only 0.04\%).
During training, we use the AdamW optimizer with a learning rate of 1e-4.
The batch size is 4, with 1050 training steps per epoch. 
For testing, we set the temperature coefficient of the large model to 0.6 and the top-p value to 0.9.
The model is implemented using PyTorch 2.3.0, and the entire training process is conducted on a single RTX 4090 GPU.

\begin{table*}[t]
\centering
\scalebox{0.82}{
\renewcommand\arraystretch{1}
\setlength{\tabcolsep}{7.9pt}{
  \begin{tabular}{@{}c|c|c|c|c|c|ccccccc@{}}
    \toprule
    \multirow{2}{*}{Methods} &
    \multicolumn{2}{c|}{Representation} & 
    \multicolumn{3}{c|}{Module Loss} &
    \multicolumn{1}{c}{\multirow{2}{*}{B1}} &\multirow{2}{*}{B2} &
    \multirow{2}{*}{B3} & \multirow{2}{*}{B4} & 
    \multirow{2}{*}{M} & \multirow{2}{*}{RG} & 
    \multirow{2}{*}{C} \\
    \cline{2-6}
    \rule{0pt}{10pt}
    & \,\,\textit{Visual}\,&\,\,\textit{Textual}\, &\,\,$\mathcal{L}_{SCA}$\, &\,\,\,$\mathcal{L}_{ECA}$\,\,&\, $\mathcal{L}_{TCA}$ \,&\multicolumn{1}{c}{} &&&&&& \\
    \midrule
    \rule{0pt}{8pt}
    Baseline &\usym{2717} &\usym{2717} & \usym{2717}& \usym{2717} &\usym{2717} &49.6 &39.9  &33.4  &28.6  &30.7  &48.4  &50.8   \\
    \midrule
    (a) &\usym{2717} &\usym{2717} & \usym{2713} &\usym{2717} & \usym{2717} &51.3 & 41.0 & 34.5 & 29.7 & 31.2 & 47.7 & 70.8   \\
    (b) &\usym{2713} &\usym{2713} & \usym{2717} &\usym{2713} & \usym{2717} &51.3  & 41.4  &35.0    &30.1   & 31.2  & 48.7  & 71.3    \\
    
    (c) & \usym{2713} & \usym{2713} &\usym{2713} &\usym{2713} &\usym{2717} &{\bfseries51.7} & 41.4 & 34.6 & 29.5 & 31.2 & 48.0 & 72.8 \\
    (d) & \usym{2713} &\usym{2713} &\usym{2717} &\usym{2713} &\usym{2713} &48.9 & 38.8 & 32.2 & 27.4 & 30.0 & 46.3 & 58.3 \\
    (e) & \usym{2713} &BERT&\usym{2713} & \usym{2713} & \usym{2713} &49.5 &39.0 &31.8 &26.3 & 30.4 & 45.4 & 48.5 \\
    \midrule
    \rule{0pt}{8pt}
    Ours& \usym{2713} &\usym{2713} & \usym{2713} & \usym{2713} &\usym{2713} &{51.5}& {\bfseries42.0}& {\bfseries35.7} & {\bfseries30.9}& {\bfseries31.4} &  {\bfseries49.0}& {\bfseries80.0}\\
    \bottomrule
  \end{tabular}
}
}
\caption{Ablation studies of our proposed method. 
The \textbf{Baseline} model is an encoder-decoder framework without an alignment mechanism.
\textbf{(a)}, \textbf{(b)}, \textbf{(c)}, \textbf{(d)} and \textbf{(e)} respectively denote the use of different representations (i.e., visual and textual in ECA and TCA) and module losses.
}
  \label{tb1:table2}
\end{table*}

\subsection{Quantitative Analysis}
We compare the proposed PCRL with some competitive brain CT report generation methods (\textbf{WGAM} \citep{yang2021weakly}, \textbf{PGCA} \citep{Shi2023Granularity}, \textbf{WGAM-HI}\citep{zhang2023weakly}). Besides, we also reproduced some SOTA models in image captioning (\textbf{HRNN} \citep{krause2017hierarchical}, \textbf{Up-Down} \citep{anderson2018bottom}) and chest X-ray report generation (\textbf{WCL} \citep{yan21weakly}, \textbf{R2Gen-CMN} \citep{chen2022cross}, \textbf{XProNet} \citep{wang2022cross}) for comprehensive comparisons on \textit{CTRG-Brain} dataset.
What's more, for fair comparisons, We also reproduce several related LLM-based methods (\textbf{LLaVA-Med}\citep{li2023llava}, \textbf{HILT}\citep{liu2024benchmarking}, \textbf{PromptMRG}\citep{Jin2024PromptMRG}with the same LLM decoder. 

As shown in Table~\ref{tb1:table1}, our method outperforms others across most evaluation metrics. 
WGAM and WGAM-HI employ weakly-supervised visual attention to extract key visual features, resulting in higher BLEU scores.
With contrastive learning for cross-modal alignment, WCL and PGCA can effectively learn relations between CT images and reports, achieving better results. However, due to the lack of training data, the above methods still produce reports that fall short in fluency and readability. Here's a refined version of your sentence:
The LLM-based methods (LLaVA-med, HILT, PromptMRG) demonstrate poorer performance compared to traditional transformer-based methods (R2Gen-CMN, XProNet). This suggests that prior knowledge from pretrained large models, without further cross-modal alignment, may lead to hallucinations in report generation.

Our PCRL achieves fine-grained cross-modal alignment by leveraging a series of pre-trained large models, generating more fluent reports and achieves the best performance compared to other methods. It is noteworthy that we tested our method using the Jieba\footnote{https://github.com/fxsjy/jieba} and LLaMA tokenizers respectively to compute NLG scores, with the latter achieving the best performance across all NLG metrics.

This discrepancy may be due to differences in tokenization methods used during training. While other models employ the traditional Jieba tool for Chinese word tokenization, our PCRL follows more advanced BPE-based subword tokenization. 
Nevertheless, our model also achieves competitive results in overall metrics.
\subsection{Ablation Study}
To evaluate the effect of each component in PCRL, we have done plenty of ablation studies, as shown in Table~\ref{tb1:table2}. Baseline is the standard encoder-decoder (ResNet101-LLaMA3) architecture without alignment. By progressively adding visual or textual representation (\textit{visual} and \textit{textual}) and the module losses ($\mathcal{L}_{SCA}$, $\mathcal{L}_{ECA}$ and $\mathcal{L}_{TCA}$), respectively denoting the incorporation of two modalities of representations and the utilization of three pathological clue-driven alignments (i.e., SCA, ECA, and TCA).

\begin{figure*}
    \centering
    \includegraphics[width=1.00\linewidth]{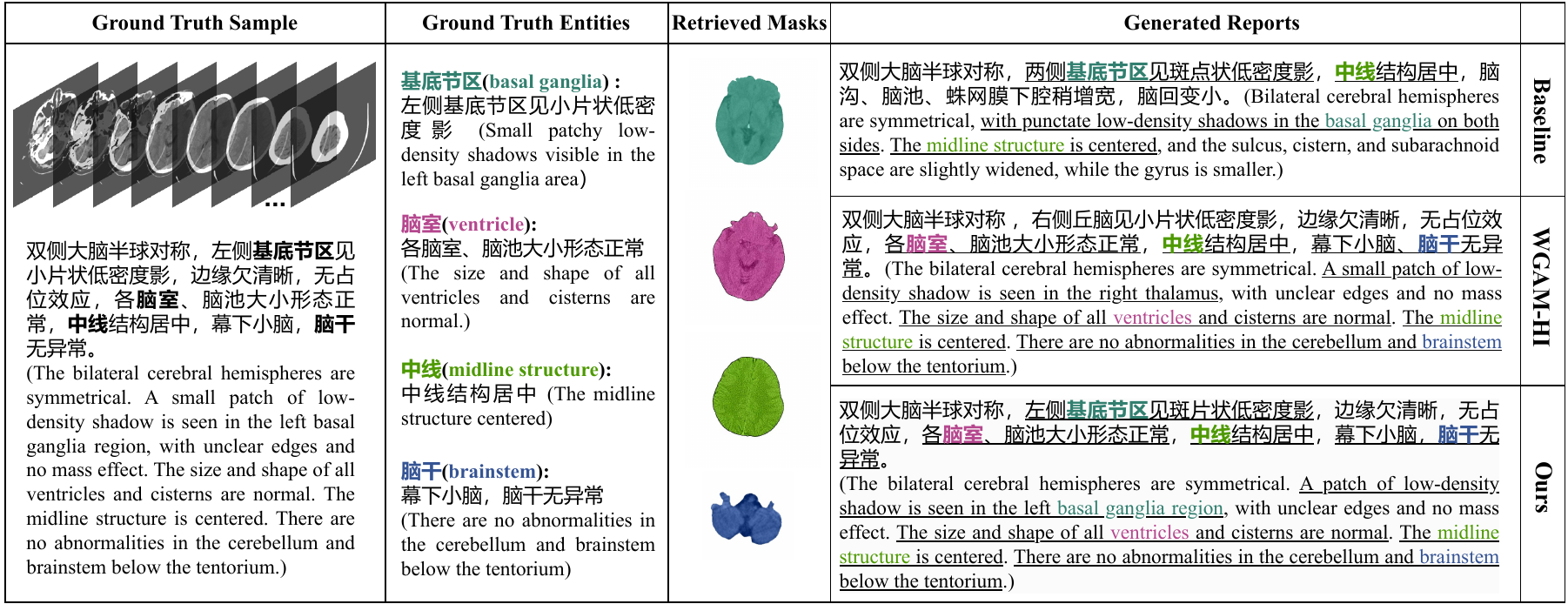}
    \caption{Visualization of report generation and mask segmentation. Given the ground truth sample and corresponding entities, the retrieved entity masks are listed in the third column. Reports generated by Baseline, WGAM-HI, and our model are listed in the fourth column. Different colors denote the specific entity words and entity masks, respectively. The English translation is given for a better understanding of the original Chinese reports in CTRG-Brain.}
    \label{fig:qualitative_report}
\end{figure*}


By comparing (a) with the baseline, we observe that incorporating the segmentation alignment significantly enhances the model's performance. This demonstrates that our SCA method effectively filters out irrelevant information from CT images and extracts crucial pathology-related visual regions.
In contrast, with the implementation of ECA, (b) achieves comparable or better performance across all metrics compared to (a). This indicates that fine-grained cross-modal pathological entity alignment can effectively learn strongly correlated visual-text representations, thus better supporting medical report generation. 
Besides, (c) combines both SCA and ECA and shows improved performance in the B1 and C metrics, which are highly correlated with keyword frequency. 
However, it slightly underperforms in the B2, B4 and RG metrics compared to (b), which is more related to overall text style.(d) showed SCA is essential for focusing on entity details and maintaining the thematic style.

Notably, to validate the contribution of joint representation learning and report generation, we using BERT language model for textual representation during alignment in (e). The result indicated that shared representation through joint learning with unified language model can effectively improve the overall quality of the task.

\subsection{Qualitative Analysis}
We visualize the brain CT reports generated by baseline, WGAM-HI~\citep{zhang2023weakly} and our model in Figure~\ref{fig:qualitative_report}.
Given the ground truth brain CT sample and entity data, our model generates better brain CT report with the most accurate entity words (e.g. basal ganglia, ventricle, and brainstem) among the competitors, which demonstrates the effectiveness of using diverse medical clues to learn enriched multi-modal representations.
It can be also seen that compared with the baseline, the report generated by our model has a better semantic structure, indicating the contribution of employing a unified LLM to transfer useful representations.
\begin{figure}
    \centering    \includegraphics[width=1.0\linewidth]{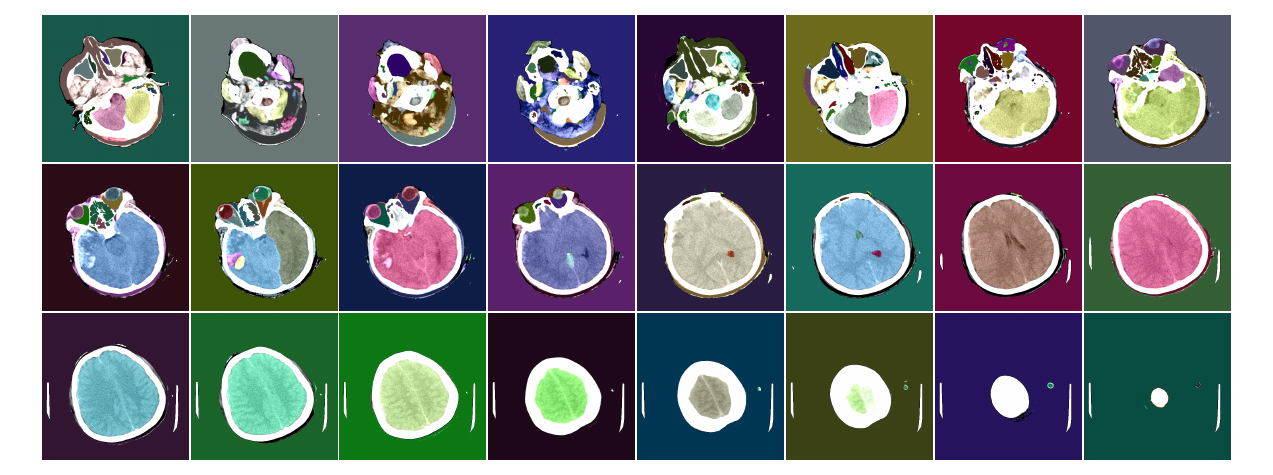}
    \caption{Visualization of all the segementation masks generated by SAM.}
    \label{fig:sam_result}
\end{figure}
Besides, we also find that the retrieved entity masks from segmentation masks gallery (see Figure ~\ref{fig:sam_result}) can generally match related entity words at both levels of visual and semantic. 
For example, the mask of ``midline structure'' and ``brainstem'' matches the empirical scan slice and fine-grained region chosen by experienced radiologists.
This guarantees the model to mine accurate visual cranial patterns, therefore generating high-quality reports.

\section{Conclusion}
We propose a novel model to mine pathological clues for enhancing multi-modal representations and seamlessly transfer them into report generation.
First, through carefully designed segmentation clue alignment, entity clue alignment, and theme clue alignment, the diverse and precise feature representation can be well-constructed.
Second, we transfer the learned representation to boost the brain CT report generation via a unified LLM prompted by task-tailored instructions.
Experiments demonstrate the effectiveness of our model in generating pathologically accurate reports.

\section*{Limitations}
Although the segmentation clues retrieved by MedClip~\citep{wang2022medclip} can generally match corresponding pathological entities and help the model neglect redundant visual information, it should be noted that a part of retrieved entity masks may not be precise. This is because MedClip is mainly pretrained by chest X-ray data with limited brain CT samples.
Thus, addressing this challenge is imperative for the research community.
In the future, we will work on exploring useful approaches.
One potential approach is to train a unified text-prompted medical segmentation model towards 3D brain CT scans, which can not only be employed to offer fine-grained visual information for medical report generation but also for other related tasks, e.g. medical VQA.

\bibliography{custom}

\begin{thebibliography}{48}
\providecommand{\natexlab}[1]{#1}

\bibitem[{Anderson et~al.(2018)Anderson, He, Buehler, Teney, Johnson, Gould, and Zhang}]{anderson2018bottom}
Peter Anderson, Xiaodong He, Chris Buehler, Damien Teney, Mark Johnson, Stephen Gould, and Lei Zhang. 2018.
\newblock Bottom-up and top-down attention for image captioning and visual question answering.
\newblock In \emph{2018 {IEEE} Conference on Computer Vision and Pattern Recognition, {CVPR} 2018, Salt Lake City, UT, USA, June 18-22, 2018}, pages 6077--6086.

\bibitem[{Banerjee and Lavie()}]{banerjeemeteor}
S~Banerjee and A~Lavie.
\newblock Meteor: an automatic metric for mt evaluation with high levels of correlation with human judgments.
\newblock \emph{Proceedings of ACL-WMT}, pages 65--72.

\bibitem[{Beltagy et~al.(2019)Beltagy, Lo, and Cohan}]{Beltagy2019SciBERT}
Iz~Beltagy, Kyle Lo, and Arman Cohan. 2019.
\newblock Scibert: {A} pretrained language model for scientific text.
\newblock In \emph{Proceedings of the 2019 Conference on Empirical Methods in Natural Language Processing and the 9th International Joint Conference on Natural Language Processing, {EMNLP-IJCNLP} 2019, Hong Kong, China, November 3-7, 2019}, pages 3613--3618.

\bibitem[{Bu et~al.(2024)Bu, Song, Li, and Dai}]{Bu2024Dynamic}
Shenshen Bu, Yujie Song, Taiji Li, and Zhiming Dai. 2024.
\newblock Dynamic knowledge prompt for chest x-ray report generation.
\newblock In \emph{Proceedings of the 2024 Joint International Conference on Computational Linguistics, Language Resources and Evaluation, {LREC/COLING}}, pages 5425--5436. {ELRA} and {ICCL}.

\bibitem[{Chen et~al.(2021)Chen, Shen, Song, and Wan}]{chen2022cross}
Zhihong Chen, Yaling Shen, Yan Song, and Xiang Wan. 2021.
\newblock Cross-modal memory networks for radiology report generation.
\newblock In \emph{Proceedings of the 59th Annual Meeting of the Association for Computational Linguistics and the 11th International Joint Conference on Natural Language Processing, {ACL/IJCNLP} 2021, (Volume 1: Long Papers), Virtual Event, August 1-6, 2021}, pages 5904--5914.

\bibitem[{Chen et~al.(2020)Chen, Song, Chang, and Wan}]{chen2020generating}
Zhihong Chen, Yan Song, Tsung{-}Hui Chang, and Xiang Wan. 2020.
\newblock Generating radiology reports via memory-driven transformer.
\newblock In \emph{Proceedings of the 2020 Conference on Empirical Methods in Natural Language Processing, {EMNLP} 2020, Online, November 16-20, 2020}, pages 1439--1449.

\bibitem[{Chen et~al.(2024)Chen, Luo, Bie, and Chen}]{Chen2024Dia}
Zhixuan Chen, Luyang Luo, Yequan Bie, and Hao Chen. 2024.
\newblock Dia-llama: Towards large language model-driven {CT} report generation.
\newblock \emph{CoRR}, abs/2403.16386.

\bibitem[{Chilamkurthy et~al.(2018)Chilamkurthy, Ghosh, Tanamala, Biviji, Campeau, Venugopal, Mahajan, Rao, and Warier}]{chilamkurthy2018development}
Sasank Chilamkurthy, Rohit Ghosh, Swetha Tanamala, Mustafa Biviji, Norbert~G. Campeau, Vasantha~Kumar Venugopal, Vidur Mahajan, Pooja Rao, and Prashant Warier. 2018.
\newblock Development and validation of deep learning algorithms for detection of critical findings in head ct scans.
\newblock \emph{arXiv preprint arXiv:1803.05854}.

\bibitem[{He et~al.(2016)He, Zhang, Ren, and Sun}]{he2016deep}
Kaiming He, Xiangyu Zhang, Shaoqing Ren, and Jian Sun. 2016.
\newblock Deep residual learning for image recognition.
\newblock In \emph{2016 {IEEE} Conference on Computer Vision and Pattern Recognition, {CVPR} 2016, Las Vegas, NV, USA, June 27-30, 2016}, pages 770--778.

\bibitem[{Hu et~al.(2021)Hu, Shen, Wallis, Allen-Zhu, Li, Wang, Wang, and Chen}]{hu2021lora}
Edward~J Hu, Yelong Shen, Phillip Wallis, Zeyuan Allen-Zhu, Yuanzhi Li, Shean Wang, Lu~Wang, and Weizhu Chen. 2021.
\newblock Lora: Low-rank adaptation of large language models.
\newblock \emph{arXiv preprint arXiv:2106.09685}.

\bibitem[{Huh et~al.(2024)Huh, Cheung, Wang, and Isola}]{huh2024platonic}
Minyoung Huh, Brian Cheung, Tongzhou Wang, and Phillip Isola. 2024.
\newblock The platonic representation hypothesis.
\newblock \emph{arXiv preprint arXiv:2405.07987}.

\bibitem[{Jin et~al.(2024)Jin, Che, Lin, and Chen}]{Jin2024PromptMRG}
Haibo Jin, Haoxuan Che, Yi~Lin, and Hao Chen. 2024.
\newblock Promptmrg: Diagnosis-driven prompts for medical report generation.
\newblock In \emph{Thirty-Eighth {AAAI} Conference on Artificial Intelligence, {AAAI}}, pages 2607--2615.

\bibitem[{Jing et~al.(2020)Jing, Wang, and Xing}]{jing2020show}
Baoyu Jing, Zeya Wang, and Eric Xing. 2020.
\newblock Show, describe and conclude: On exploiting the structure information of chest x-ray reports.
\newblock \emph{arXiv preprint arXiv:2004.12274}.

\bibitem[{Jing et~al.(2018)Jing, Xie, and Xing}]{jing2018automatic}
Baoyu Jing, Pengtao Xie, and Eric~P. Xing. 2018.
\newblock On the automatic generation of medical imaging reports.
\newblock In \emph{Proceedings of the 56th Annual Meeting of the Association for Computational Linguistics, 2018, Melbourne, Australia, July 15-20, 2018, Volume 1: Long Papers}, pages 2577--2586.

\bibitem[{Kirillov et~al.(2023)Kirillov, Mintun, Ravi, Mao, Rolland, Gustafson, Xiao, Whitehead, Berg, Lo et~al.}]{kirillov2023segment}
Alexander Kirillov, Eric Mintun, Nikhila Ravi, Hanzi Mao, Chloe Rolland, Laura Gustafson, Tete Xiao, Spencer Whitehead, Alexander~C Berg, Wan-Yen Lo, et~al. 2023.
\newblock Segment anything.
\newblock In \emph{Proceedings of the IEEE/CVF International Conference on Computer Vision}, pages 4015--4026.

\bibitem[{Krause et~al.(2017)Krause, Johnson, Krishna, and Fei{-}Fei}]{krause2017hierarchical}
Jonathan Krause, Justin Johnson, Ranjay Krishna, and Li~Fei{-}Fei. 2017.
\newblock A hierarchical approach for generating descriptive image paragraphs.
\newblock In \emph{2017 {IEEE} Conference on Computer Vision and Pattern Recognition, {CVPR} 2017, Honolulu, HI, USA, July 21-26, 2017}, pages 3337--3345.

\bibitem[{Li et~al.(2023{\natexlab{a}})Li, Wong, Zhang et~al.}]{li2023llava}
C~Li, C~Wong, S~Zhang, et~al. 2023{\natexlab{a}}.
\newblock Llava-med: training a large language-and-vision assistant for biomedicine in one day. arxiv.
\newblock \emph{arXiv preprint arXiv:2306.00890}.

\bibitem[{Li et~al.(2019)Li, Liang, Hu, and Xing}]{li2019knowledge}
Christy~Y. Li, Xiaodan Liang, Zhiting Hu, and Eric~P. Xing. 2019.
\newblock Knowledge-driven encode, retrieve, paraphrase for medical image report generation.
\newblock In \emph{The Thirty-Third {AAAI} Conference on Artificial Intelligence, {AAAI} 2019, The Thirty-First Innovative Applications of Artificial Intelligence Conference, {IAAI} 2019, The Ninth {AAAI} Symposium on Educational Advances in Artificial Intelligence, {EAAI} 2019, Honolulu, Hawaii, USA, January 27 - February 1, 2019}, pages 6666--6673.

\bibitem[{Li et~al.(2023{\natexlab{b}})Li, Lin, Chen, Lin, Liang, and Chang}]{li2023dynamic}
Mingjie Li, Bingqian Lin, Zicong Chen, Haokun Lin, Xiaodan Liang, and Xiaojun Chang. 2023{\natexlab{b}}.
\newblock Dynamic graph enhanced contrastive learning for chest x-ray report generation.
\newblock In \emph{{IEEE} Conference on Computer Vision and Pattern Recognition}, pages 3334--3343.

\bibitem[{Li et~al.(2018)Li, Liang, Hu, and Xing}]{li2018hybrid}
Yuan Li, Xiaodan Liang, Zhiting Hu, and Eric~P Xing. 2018.
\newblock Hybrid retrieval-generation reinforced agent for medical image report generation.
\newblock \emph{Advances in neural information processing systems}, 31.

\bibitem[{Lin(2004)}]{lin2004rouge}
Chin-Yew Lin. 2004.
\newblock Rouge: A package for automatic evaluation of summaries.
\newblock In \emph{Text summarization branches out}, pages 74--81.

\bibitem[{Liu et~al.(2024)Liu, Wan, Wang, Shen, Wang, Zheng, Zhang, and Arcucci}]{liu2024benchmarking}
Che Liu, Zhongwei Wan, Yuqi Wang, Hui Shen, Haozhe Wang, Kangyu Zheng, Mi~Zhang, and Rossella Arcucci. 2024.
\newblock Benchmarking and boosting radiology report generation for 3d high-resolution medical images.
\newblock \emph{arXiv preprint arXiv:2406.07146}.

\bibitem[{Liu et~al.(2021{\natexlab{a}})Liu, Ge, and Wu}]{liu2021competence}
Fenglin Liu, Shen Ge, and Xian Wu. 2021{\natexlab{a}}.
\newblock Competence-based multimodal curriculum learning for medical report generation.
\newblock In \emph{Proceedings of the 59th Annual Meeting of the Association for Computational Linguistics and the 11th International Joint Conference on Natural Language Processing, {ACL/IJCNLP} 2021, (Volume 1: Long Papers), Virtual Event, August 1-6, 2021}, pages 3001--3012.

\bibitem[{Liu et~al.(2021{\natexlab{b}})Liu, Wu, Ge, Fan, and Zou}]{liu2021exploring}
Fenglin Liu, Xian Wu, Shen Ge, Wei Fan, and Yuexian Zou. 2021{\natexlab{b}}.
\newblock Exploring and distilling posterior and prior knowledge for radiology report generation.
\newblock In \emph{{IEEE} Conference on Computer Vision and Pattern Recognition, {CVPR} 2021, virtual, June 19-25, 2021}, pages 13753--13762.

\bibitem[{Liu et~al.(2021{\natexlab{c}})Liu, Yin, Wu, Ge, Zou, Zhang, and Sun}]{liu2021contrastive}
Fenglin Liu, Changchang Yin, Xian Wu, Shen Ge, Yuexian Zou, Ping Zhang, and Xu~Sun. 2021{\natexlab{c}}.
\newblock Contrastive attention for automatic chest x-ray report generation.
\newblock \emph{arXiv preprint arXiv:2106.06965}.

\bibitem[{Meta(2024)}]{meta2024introducing}
AI~Meta. 2024.
\newblock Introducing meta llama 3: The most capable openly available llm to date.
\newblock \emph{Meta AI.}

\bibitem[{Papineni et~al.(2002)Papineni, Roukos, Ward, and Zhu}]{papineni2002bleu}
Kishore Papineni, Salim Roukos, Todd Ward, and Wei{-}Jing Zhu. 2002.
\newblock Bleu: a method for automatic evaluation of machine translation.
\newblock In \emph{Proceedings of the 40th Annual Meeting of the Association for Computational Linguistics, July 6-12, 2002, Philadelphia, PA, {USA}}, pages 311--318.

\bibitem[{Radford et~al.(2021)Radford, Kim, Hallacy, Ramesh, Goh, Agarwal, Sastry, Askell, Mishkin, Clark, Krueger, and Sutskever}]{radford2021learning}
Alec Radford, Jong~Wook Kim, Chris Hallacy, Aditya Ramesh, Gabriel Goh, Sandhini Agarwal, Girish Sastry, Amanda Askell, Pamela Mishkin, Jack Clark, Gretchen Krueger, and Ilya Sutskever. 2021.
\newblock Learning transferable visual models from natural language supervision.
\newblock In \emph{Proceedings of the 38th International Conference on Machine Learning, {ICML} 2021, 18-24 July 2021, Virtual Event}, pages 8748--8763.

\bibitem[{Shen et~al.(2024)Shen, Shi, Zhang, Ji, Liu, and Xu}]{shen2024ghcl}
Qingya Shen, Yanzhao Shi, Xiaodan Zhang, Junzhong Ji, Ying Liu, and Huimin Xu. 2024.
\newblock Ghcl: Gaussian heuristic curriculum learning for brain ct report generation.
\newblock \emph{Multimedia Systems}, 30(2):69.

\bibitem[{Shi et~al.(2024)Shi, Ji, Zhang, Liu, Wang, and Xu}]{Shi2024Prior}
Yanzhao Shi, Junzhong Ji, Xiaodan Zhang, Ying Liu, Zheng Wang, and Huimin Xu. 2024.
\newblock Prior tissue knowledge-driven contrastive learning for brain {CT} report generation.
\newblock \emph{Multim. Syst.}, 30(2):98.

\bibitem[{Shi et~al.(2023)Shi, Ji, Zhang, Qu, and Liu}]{Shi2023Granularity}
Yanzhao Shi, Junzhong Ji, Xiaodan Zhang, Liangqiong Qu, and Ying Liu. 2023.
\newblock Granularity matters: Pathological graph-driven cross-modal alignment for brain {CT} report generation.
\newblock In \emph{Proceedings of the 2023 Conference on Empirical Methods in Natural Language Processing, {EMNLP}}, pages 6617--6630.

\bibitem[{Tang et~al.(2024)Tang, Yang, Zhang, and Yuan}]{tang2024work}
Yuhao Tang, Haichen Yang, Liyan Zhang, and Ye~Yuan. 2024.
\newblock Work like a doctor: Unifying scan localizer and dynamic generator for automated computed tomography report generation.
\newblock \emph{Expert Systems with Applications}, 237:121442.

\bibitem[{Thawakar et~al.(2023)Thawakar, Shaker, Mullappilly, Cholakkal, Anwer, Khan, Laaksonen, and Khan}]{Thawakar2023XrayGPT}
Omkar Thawakar, Abdelrahman~M. Shaker, Sahal~Shaji Mullappilly, Hisham Cholakkal, Rao~Muhammad Anwer, Salman~H. Khan, Jorma Laaksonen, and Fahad~Shahbaz Khan. 2023.
\newblock Xraygpt: Chest radiographs summarization using medical vision-language models.
\newblock \emph{CoRR}, abs/2306.07971.

\bibitem[{van~den Oord et~al.(2018)van~den Oord, Li, and Vinyals}]{oord2018representation}
A{\"{a}}ron van~den Oord, Yazhe Li, and Oriol Vinyals. 2018.
\newblock Representation learning with contrastive predictive coding.
\newblock \emph{arXiv preprint arXiv:1807.03748}.

\bibitem[{Vedantam et~al.(2015)Vedantam, Zitnick, and Parikh}]{vedantam2015cider}
Ramakrishna Vedantam, C.~Lawrence Zitnick, and Devi Parikh. 2015.
\newblock Cider: Consensus-based image description evaluation.
\newblock In \emph{{IEEE} Conference on Computer Vision and Pattern Recognition, {CVPR} 2015, Boston, MA, USA, June 7-12, 2015}, pages 4566--4575.

\bibitem[{Vinyals et~al.(2015)Vinyals, Toshev, Bengio, and Erhan}]{vinyals2015show}
Oriol Vinyals, Alexander Toshev, Samy Bengio, and Dumitru Erhan. 2015.
\newblock Show and tell: {A} neural image caption generator.
\newblock In \emph{{IEEE} Conference on Computer Vision and Pattern Recognition, {CVPR} 2015, Boston, MA, USA, June 7-12, 2015}, pages 3156--3164.

\bibitem[{Wang et~al.(2022{\natexlab{a}})Wang, Bhalerao, and He}]{wang2022cross}
Jun Wang, Abhir Bhalerao, and Yulan He. 2022{\natexlab{a}}.
\newblock Cross-modal prototype driven network for radiology report generation.
\newblock In \emph{Computer Vision - {ECCV} 2022 - 17th European Conference, Tel Aviv, Israel, October 23-27, 2022, Proceedings, Part {XXXV}}, pages 563--579. Springer.

\bibitem[{Wang et~al.(2023)Wang, Yang, Huang, Yang, Majumder, and Wei}]{Wang2023ImprovingTE}
Liang Wang, Nan Yang, Xiaolong Huang, Linjun Yang, Rangan Majumder, and Furu Wei. 2023.
\newblock Improving text embeddings with large language models.
\newblock \emph{ArXiv}, abs/2401.00368.

\bibitem[{Wang et~al.(2018)Wang, Peng, Lu, Lu, and Summers}]{wang2018tienet}
Xiaosong Wang, Yifan Peng, Le~Lu, Zhiyong Lu, and Ronald~M. Summers. 2018.
\newblock Tienet: Text-image embedding network for common thorax disease classification and reporting in chest x-rays.
\newblock In \emph{2018 {IEEE} Conference on Computer Vision and Pattern Recognition, {CVPR} 2018, Salt Lake City, UT, USA, June 18-22, 2018}, pages 9049--9058.

\bibitem[{Wang et~al.(2022{\natexlab{b}})Wang, Wu, Agarwal, and Sun}]{wang2022medclip}
Zifeng Wang, Zhenbang Wu, Dinesh Agarwal, and Jimeng Sun. 2022{\natexlab{b}}.
\newblock Medclip: Contrastive learning from unpaired medical images and text.
\newblock \emph{arXiv preprint arXiv:2210.10163}.

\bibitem[{Xu et~al.(2015)Xu, Ba, Kiros, Cho, Courville, Salakhutdinov, Zemel, and Bengio}]{xu2015show}
Kelvin Xu, Jimmy Ba, Ryan Kiros, Kyunghyun Cho, Aaron~C. Courville, Ruslan Salakhutdinov, Richard~S. Zemel, and Yoshua Bengio. 2015.
\newblock Show, attend and tell: Neural image caption generation with visual attention.
\newblock In \emph{Proceedings of the 32nd International Conference on Machine Learning, {ICML} 2015, Lille, France, 6-11 July 2015}, pages 2048--2057.

\bibitem[{Yan et~al.(2021)Yan, He, Lu, Du, Chang, Gentili, McAuley, and Hsu}]{yan21weakly}
An~Yan, Zexue He, Xing Lu, Jiang Du, Eric~Y. Chang, Amilcare Gentili, Julian~J. McAuley, and Chun{-}Nan Hsu. 2021.
\newblock Weakly supervised contrastive learning for chest x-ray report generation.
\newblock In \emph{Findings of the Association for Computational Linguistics: {EMNLP} 2021, Virtual Event / Punta Cana, Dominican Republic, 16-20 November, 2021}, pages 4009--4015.

\bibitem[{Yang et~al.(2022)Yang, Wu, Ge, Zhou, and Xiao}]{yang2022knowledge}
Shuxin Yang, Xian Wu, Shen Ge, S~Kevin Zhou, and Li~Xiao. 2022.
\newblock Knowledge matters: Chest radiology report generation with general and specific knowledge.
\newblock \emph{Medical image analysis}, 80:102510.

\bibitem[{Yang et~al.(2021)Yang, Ji, Zhang, Liu, and Wang}]{yang2021weakly}
Sisi Yang, Junzhong Ji, Xiaodan Zhang, Ying Liu, and Zheng Wang. 2021.
\newblock Weakly guided hierarchical encoder-decoder network for brain ct report generation.
\newblock In \emph{{IEEE} International Conference on Bioinformatics and Biomedicine, {BIBM} 2021, Houston, TX, USA, December 9-12, 2021}, pages 568--573. IEEE.

\bibitem[{You et~al.(2021)You, Liu, Ge, Xie, Zhang, and Wu}]{you2021aligntransformer}
Di~You, Fenglin Liu, Shen Ge, Xiaoxia Xie, Jing Zhang, and Xian Wu. 2021.
\newblock Aligntransformer: Hierarchical alignment of visual regions and disease tags for medical report generation.
\newblock In \emph{Medical Image Computing and Computer Assisted Intervention - {MICCAI} 2021 - 24th International Conference, Strasbourg, France, September 27 - October 1, 2021, Proceedings, Part {III}}, pages 72--82.

\bibitem[{Yu et~al.(2023)Yu, Lezama, Gundavarapu, Versari, Sohn, Minnen, Cheng, Gupta, Gu, Hauptmann et~al.}]{yu2023language}
Lijun Yu, Jos{\'e} Lezama, Nitesh~B Gundavarapu, Luca Versari, Kihyuk Sohn, David Minnen, Yong Cheng, Agrim Gupta, Xiuye Gu, Alexander~G Hauptmann, et~al. 2023.
\newblock Language model beats diffusion--tokenizer is key to visual generation.
\newblock \emph{arXiv preprint arXiv:2310.05737}.

\bibitem[{Zhang et~al.(2023)Zhang, Yang, Shi, Ji, Liu, Wang, and Xu}]{zhang2023weakly}
Xiaodan Zhang, Sisi Yang, Yanzhao Shi, Junzhong Ji, Ying Liu, Zheng Wang, and Huimin Xu. 2023.
\newblock Weakly guided attention model with hierarchical interaction for brain ct report generation.
\newblock \emph{Computers in Biology and Medicine}, page 107650.

\bibitem[{Zhang et~al.(2020)Zhang, Wang, Xu, Yu, Yuille, and Xu}]{zhang2020radiology}
Yixiao Zhang, Xiaosong Wang, Ziyue Xu, Qihang Yu, Alan~L. Yuille, and Daguang Xu. 2020.
\newblock When radiology report generation meets knowledge graph.
\newblock In \emph{The Thirty-Fourth {AAAI} Conference on Artificial Intelligence, {AAAI} 2020, The Thirty-Second Innovative Applications of Artificial Intelligence Conference, {IAAI} 2020, The Tenth {AAAI} Symposium on Educational Advances in Artificial Intelligence, {EAAI} 2020, New York, NY, USA, February 7-12, 2020}, pages 12910--12917.

\end{thebibliography}
\end{document}